\documentclass[letterpaper, 10 pt, conference]{ieeeconf}  

\usepackage{amsmath}
\usepackage{amssymb} 

\renewcommand{\QED}{\QEDopen}

\usepackage[labelformat=simple]{subcaption}
\DeclareCaptionLabelSeparator{periodspace}{.\quad}
\usepackage{tikz}
\usetikzlibrary{arrows,matrix,positioning,patterns}
\usetikzlibrary{calc}
\usepackage{pgfplots} 
\usetikzlibrary{plotmarks}

\usepackage[utf8]{inputenc}
\usepackage{pgfplots}
\usepgfplotslibrary{groupplots}

\usepackage{url}
\usepackage{multirow}
\usepackage{colortbl}

\usepackage[ruled, vlined, linesnumbered]{algorithm2e}

\DeclareSymbolFont{bbold}{U}{bbold}{m}{n}
\DeclareSymbolFontAlphabet{\mathbbold}{bbold}

\usepackage[labelformat=simple]{subcaption}
\DeclareCaptionLabelSeparator{periodspace}{.\quad}
\usepackage{listings}
\usepackage{xcolor}
\usepackage{cite}

\usepackage{wrapfig}
\usepackage{supertabular}

\IEEEoverridecommandlockouts                              
\title{\LARGE \bf
A posteriori Trading-inspired Model-free Time Series Segmentation}

\author{Mogens Graf Plessen
\thanks{Independent research, {\tt\small mgplessen@gmail.com}}}
\renewcommand{\baselinestretch}{0.97}
\begin{document}

\maketitle
\thispagestyle{empty}
\pagestyle{empty}

\begin{abstract}
Within the context of multivariate time series segmentation this paper proposes a method inspired by a posteriori optimal trading. After a normalization step time series are treated channel-wise as surrogate stock prices that can be traded optimally a posteriori in a virtual portfolio holding either stock or cash. Linear transaction costs are interpreted as hyperparameters for noise filtering. Resulting trading signals as well as resulting trading signals obtained on the reversed time series are used for unsupervised labeling, before a consensus over channels is reached that determines segmentation time instants. The method is model-free such that no model prescriptions for segments are made. Benefits of proposed approach include simplicity, adaptability to a wide range of different shapes of time series, and in particular computational efficiency that make it suitable for big data. Performance is demonstrated on synthetic and real-world data, including a large-scale dataset comprising a multivariate time series of dimension 1000 and length 2709. Proposed method is compared to a popular model-based bottom-up approach fitting piecewise affine models and to a state-of-the-art model-based top-down approach fitting Gaussian models, and found to be consistently faster while producing more intuitive results.
\end{abstract}

\section{Introduction\label{sec_intro}}

Analysis of multivariate time series data is relevant in every engineering field. An ongoing increase in sensors employment simultaneously implies a rise in measurement data generation. Once a multivariate data point is measured it can either be processed isolatedly or in combination with a sequence of previous measurements. Given such a sequence a natural task is to segment it. The problem addressed in this paper is to partition multivariate time series data into segments such that different segments exhibit different and characteristic behavior. 

The importance of time series segmentation stems from the fact that it is essential to manage large amounts of multivariate data, and that it can form the foundation for further upstream time series analysis tasks such as clustering, compression or forecasting. Note that the problem of time series segmentation is closely related to, and often interchangeably implied by a variety of problems labeled in the literature as ``changepoint detection'', ``breakpoint detection'' or ``event detection''. For example, segment boundaries can be interpreted as changepoints or breakpoints where characteristic behavior changes.

\newlength\figureheight
\newlength\figurewidth
\setlength\figureheight{5cm}
\setlength\figurewidth{5cm}
\begin{figure}
\centering
\input{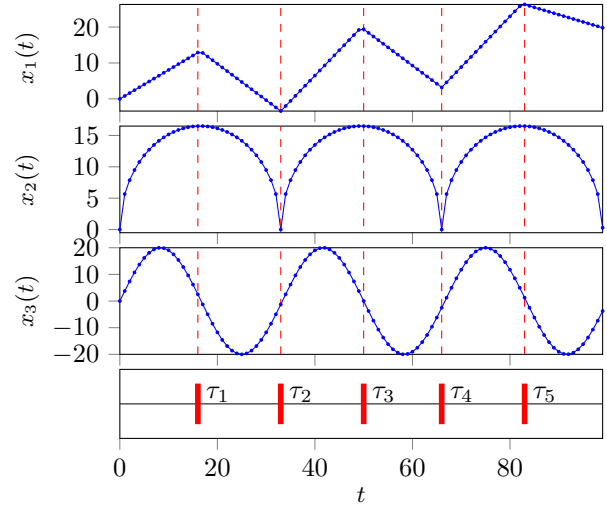}
\caption{Problem sketch. Given a multivariate time series (blue) suitable segmentation time instants (red dashed) are sought. In this paper, a method inspired by a posteriori optimal trading is proposed that identifies local maxima and minima channel-wise for each $\{x_i(t)\}_{t=0}^T,\forall i=1,\dots,n_x$ and for their reversed time series, before a unifying consensus over all channels is reached resulting in time instants $\{\tau_k\}_{k=1}^{\hat{K}}$. For example, for $n_x=3$ and a uniformly-weighted consensus over all three channels the result is obtained as illustrated, eventhough the third channel's local maxima and minima are phase-shifted with respect to the other two channels.}
\label{fig_figProblSketch}
\end{figure}

\begin{table}
\vspace{-0.4cm}
\centering
\begin{tabular}{|ll|}
\hline
\multicolumn{2}{|c|}{MAIN NOMENCLATURE}\\
\multicolumn{2}{|l|}{Symbols}\\
$b_i(t)$ & Binary state for channel $i$ at time $t$.\\
$b_i^\star(t)$ & Optimal binary state for channel $i$ at time $t$.\\
$\epsilon_i$ & Transaction cost level for channel $i$.\\
$\epsilon^\text{min},\epsilon^\text{max}$ & Minimum and maximum constraint on $\epsilon_i$.\\
$\gamma^{\bullet}$ & Hyperparameters for APTS, $\bullet\in\{\text{mult},\text{close},\text{plat}\}$.\\
$n_x$ & Dimension of multivariate time series.\\
$q^\star(t)$ & Optimal binary consensus state at time $t$.\\
$\{\tau_k\}_{k=1}^{\hat{K}}$ & Identified list of segmentation time instants.\\
$\{\tau_k^\text{flip}\}_{k=1}^{\hat{K}^\text{flip}}$ & Segmentation obtained on reversed time series.\\
$u_i(t)$ & Binary control for channel $i$ at time $t$.\\
$x(t)$ & Multivariate time series data point at time $t$.\\
$x_i(t)$ & Data point for channel $i$ at time $t$.\\
$\tilde{x}_i(t)$ & Data point after normalization.\\
$z_i(t)$ & Four-dimensional state vector.\\
$\hat{K}$ & Identified number of segments.\\
$K^\text{max}$ & Maximum permissible number of segments.\\
$T$ & Length of multivariate time series.\\[3pt]
\multicolumn{2}{|l|}{Functions}\\
$\mathcal{F}^\text{normalize}(\cdot)$ & Mapping from $\{x_i(t)\}_{t=0}^{T}$ to $\{\tilde{x}_i(t)\}_{t=0}^{T}$.\\
$\mathcal{F}^\text{trade}(\cdot)$ & Mapping from $\{\tilde{x}_i(t)\}_{t=0}^{T}$ to $\{b_i^\star(t)\}_{t=0}^{T}$.\\
$\mathcal{F}^\text{consensus}(\cdot)$ & Mapping from $\{\{b_i^\star(t)\}_{t=0}^{T}\}_{i=1}^{n_x}$ to $\{q^\star(t)\}_{t=0}^{T}$.\\
$\mathcal{F}^\text{terminate}(\cdot)$ & Mapping from $\{b_i^\star(t)\}_{t=0}^{T}$ and $K^\text{max}$ to a binary.\\
$\mathcal{F}^\text{merge}(\cdot)$ & Mapping for merging of $\{ \tau_k \}_{k=1}^{\hat{K}}$ and $\{ \tau_k^\text{flip} \}_{k=1}^{\hat{K}^\text{flip}}$.\\
$\mathcal{F}^\text{epsilon}(\cdot)$ & Mapping for determining transaction cost level $\epsilon_i$.\\[3pt]
\multicolumn{2}{|l|}{Abbreviations}\\
BU & Bottom-Up Algorithm \cite{keogh2001online}. \\
GGS & Greedy Gaussian Segmentation Algorithm \cite{hallac2019greedy}. \\
APTS & A Posteriori Trading-inspired Segmentation. \\
\hline
\end{tabular}
\end{table}

Algorithms for time series segmentation can be classified according to four criteria. First, the \emph{number of segments} is either prescribed by a hyperparameter choice, or, alternatively, optimized in addition to the time instants defining the segmentation. Second, each segment may be fitted by a specific \emph{segment model} (e.g., a piecewise affine or Gaussian model), or, alternatively, no specific model fit may be prescribed. For the former case, it can be further differentiated between the cases of (i) fitting the same model class to all segments but with different parameters, or (ii) fitting each segment by a model selected from a set of different model classes. Third, for the \emph{algorithm type} it can be differentiated between batch offline or recursive online computation. Fourth, for the \emph{machine learning type} it can be distinguished between supervised and unsupervised methods. For surveys on the topic see \cite{keogh2004segmenting,aminikhanghahi2017survey, truong2019selective}. Supervised methods require a training phase to learn a classifier \cite{desobry2005online, feuz2014automated, zheng2008learning}. For example, a binary classifier distinguishes a ``transition state'' (changepoint) and a ``within-state''. See also \cite{chung2004evolutionary} for evolutionary time series segmentation using pattern templates. In contrast, unsupervised methods deal with unlabeled data. There exists a large variety of approaches, ranging from subspace  \cite{kawahara2007change}, to probabilistic \cite{adams2007bayesian}, segment-fitting \cite{fuchs2010online}, kernel-based \cite{harchaoui2009regularized}, deep learning-based \cite{lee2018time}, graph-based \cite{chen2015graph} and density ratio methods \cite{liu2013change, kawahara2012sequential}. For a method assuming a fixed number of segments see, e.g., \cite{himberg2001time}. A variety of different segment model classes have been proposed, including also a model-free approach working with a dissimilarity matrix for single pattern change detection within the remote sensing domain \cite{garg2011model}, Fourier transforms \cite{agrawal1993efficient}, wavelets \cite{sharifzadeh2005change}, piecewise affine representations \cite{keogh2001online, liu2008novel}, and multivariate Gaussians \cite{hallac2019greedy}. 

Within this context the motivation and contribution of this paper is to present a simple and fast method that (i) does not assume a fixed number of changepoints, (ii) does not assume any specific segment model, (iii) is used for batch computation, (iv) is an unsupervised algorithm, (v) parallelizable, (vi) library-free in the sense that no specific internal optimisation routine (such as, e.g., a least-squares solution or factorizaton) is required, (vii) exhibits simple-to-select hyperparameters, and (viii) is scalable to large multivariate time series data. The fundamental approach is based on ideas from a posteriori optimal trading \cite{plessen2018posteriori}. After a normalization step time series are treated channel-wise as surrogate stock prices that can be traded optimally a posteriori in a virtual portfolio holding either stock or cash. Linear transaction costs are therefore interpreted as hyperparameters for noise filtering. Resulting trading signals as well as resulting trading signals obtained on the reversed time series are then used for unsupervised labeling, before a consensus over channels is reached that determines segmentation instants. To the author's knowledge this approach has not been presented in the literature. 

The remaining paper is organized as follows. The multivariate time series segmentation problem is formulated mathematically in \S \ref{sec_problFormulation}. The proposed solution is presented in \S \ref{sec_problem_soln}. Numerical results are evaluated in \S \ref{sec_expts} and compared to a bottom-up method \cite{keogh2001online} and to a recent top-down Gaussians-fitting method \cite{hallac2019greedy} on synthetic as well as real-world datasets \cite{dau2019ucr}, before concluding with \S \ref{sec_concl}.

\section{Problem Formulation\label{sec_problFormulation}}

A multivariate time series of finite length is defined as $x(t)\in\mathbb{R}^{n_x},~\forall t=0,\dots,T$. Here, $x(T)\in\mathbb{R}^{n_x}$ represents the data point measured last. Time-indices do not need to be uniformly spaced. Instead, data points are only assumed to be ordered. Indices do not necessarily need to represent time instants, but could also denote space indices (e.g., with space along a road as the dependent variable), unitless indices or other indexing. In general, units can vary channel-wise, whereby notation $x_i(t),~\forall i=1,\dots,n_x$ is employed.

The problem addressed is to partition the multivariate time series, $\{x_i(t)\}_{t=0}^T,~\forall i=1,\dots,n_x$, into $\hat{K}+1$ segments defined by changepoints or ordered time instants, $\{\tau_k\}_{k=1}^{\hat{K}}$, with $0<\tau_1<\dots<\tau_{\hat{K}}<T$  such that different segments exhibit different and characteristic behavior. Rather than constraining a fixed number of segments, $\hat{K}$ shall here also be identified in addition to the segmenting time instants. A batch offline algorithm is developed. In this  paper, no specific model shall be prescribed for any time series segment to not artificially limit the solution space. This is in contrast to model-based approaches such as, e.g., \cite{keogh2001online} and \cite{hallac2019greedy}, where piecewise affine and Gaussian models are fitted as part of their segmentation logics, respectively.

\section{Problem Solution\label{sec_problem_soln}}

\subsection{A Posteriori Trading-inspired Segmentation Algorithm\label{subsec_ProposedSoln_offline}}

This paper builds on \cite{plessen2018posteriori} where a posteriori optimal trading over multiple assets subject to constraints for diversification was discussed. Within the context of multivariate time series segmentation the proposed method can be interpreted as virtual or surrogate channel-wise optimal trading on normalized time series data and the reversed time series data subject to linear transaction costs that represent hyperparameters for noise filtering, before determining a consensus based on the trading signals for all channels to obtain $\hat{K}$ and $\{\tau_k \}_{k=1}^{\hat{K}}$. Details are outlined in the following.

First, a normalization is carried out in order to obtain channel-wise positive time series data. As will be further discussed, positivity is a prerequisite for proposed segmentation logic. The following linear transformation is employed, $\tilde{x}_i(t) = x_i(t) + \tilde{x}_i^\text{offset},\forall t=0,\dots,T,\forall i=1,\dots,n_x$, with $\tilde{x}_i^\text{offset}=| \min_{t\in\{0,\dots,T\}} x_i(t) |+1$. The bias ensures positivity. In the following, the mapping from $\{ x_i(t) \}_{t=0}^T$ to $\{ \tilde{x}_i(t) \}_{t=0}^T$ shall be abbreviated by $\mathcal{F}^\text{normalize}( \{ x_i(t) \}_{t=0}^T )$. 

Second, channel-wise surrogate wealth dynamics are introduced that model a virtual portfolio holding either a virtual cash position or the ``stock'' modeled by the channel-wise normalized time series. A four-dimensional state vector is defined, $z_i(t)=[n_i(t),~c_i(t),~b_i(t),~w_i(t)]$, with $n_i(t)\geq 0$  the number of shares held at time $t$, $c_i(t)\geq 0$ the cash position, $b_i(t)\in\{-1,1\}$ a binary state indicating full investment in cash or stock, and finally $w_i(t)\geq 0$ denoting total wealth at time $t$. In contrast to regular stock trading for our surrogate setup the number of shares, $n_i(t)\geq 0$, is real-valued. At $t=0$ the state vector is initialized with $z_i(0)=[0,~\frac{\tilde{x}_i(0)}{1-\epsilon_i},~-1,~\frac{\tilde{x}_i(0)}{1-\epsilon_i}]$, where $\epsilon_i\in[0,1)$ is interpreted as a linear transaction cost. This initialization implies an initial cash position sufficient to buy one share when accounting for transaction cost and is defined this way to avoid adding new hyperparameters. Introducing a control variable, $u_i(t)\in\{-1,1\}$, state transition dynamics are,
\begin{equation}
z_i(t+1) = \begin{cases} 
\begin{bmatrix} 0\\ c_i(t)\\-1\\c_i(t)\end{bmatrix}, & ~\text{if}~u_i(t)=-1,\\
\begin{bmatrix}\frac{c_i(t)(1-\epsilon_i)}{\tilde{x}_i(t)}\\0\\1\\n_i(t+1)\tilde{x}_i(t+1)\end{bmatrix}, & ~\text{if}~u_i(t)=1,
\end{cases}
\label{eq_def_zitp1_a}
\end{equation}
for $z_i(t)\in\{z_i(t)\in\mathbb{R}^4:b_i(t)=-1\}$, and
\begin{equation}
z_i(t+1) = \begin{cases} 
\begin{bmatrix}0\\n_i(t)\tilde{x}_i(t)(1-\epsilon_i)\\-1\\c_i(t+1)\end{bmatrix}, & ~\text{if}~u_i(t)=-1,\\
\begin{bmatrix}n_i(t)\\0\\1\\n_i(t)\tilde{x}_i(t+1)\end{bmatrix}, & ~\text{if}~u_i(t)=1,
\end{cases}
\label{eq_def_zitp1_b}
\end{equation}
for $z_i(t)\in\{z_i(t)\in\mathbb{R}^4:b_i(t)=1\}$. Combinedly, \eqref{eq_def_zitp1_a}-\eqref{eq_def_zitp1_b} model all four possible transitions between full cash and full stock investment subject to linear transaction costs.

Third, in a causal setting $\tilde{x}_i(t+1)$ is not known at time $t$. However, in the batch setting it is available, which is equivalent to perfect one step-ahead knowledge a posteriori. Therefore, and due to the fact of a positive $\tilde{x}_i(t)$ by above discussion there always exists a wealth-maximizing trading trajectory from $t=0$ to $T$ as a function of transaction cost level $\epsilon_i\geq 0$ such that $w_i(T)$ is channel-wise maximized. This trajectory can be computed efficiently as follows. 

Starting from $z_i(0)$ defined above, at $t=1$ two possible states can result which differ by binary $b_i(1)\in\{-1,1\}$. Let these two states be denoted by $z_i^{(-1)}(1)$ and $z_i^{(1)}(1)$, respectively. Then, by Bellman's principle of optimality\cite{bellman1954theory} and with the purpose of deriving the wealth-maximizing trading trajectory the following recursion can be implemented for all $t\geq 1$. When $\tilde{x}_i(t+1)$ becomes available, $z_i^{(-1)}(t)$ and $z_i^{(1)}(t)$ branch out to a total of four different states according to \eqref{eq_def_zitp1_a}-\eqref{eq_def_zitp1_b}. These are pruned to two by selecting the $w_i(t+1)$-maximizing solutions for each $b_i(t+1)=-1$ and $b_i(t+1)=1$ such that $z_i^{(-1)}(t+1)$ and $z_i^{(1)}(t+1)$ are obtained, respectively. Their corresponding optimal parent states are further recorded, which shall be denoted by $z_i^{(-1),\text{parent}}(t)$ and $z_i^{(1),\text{parent}}(t)$. This recursion is repeated until $t=T$. Given $z_i^{(-1)}(T)$ and $z_i^{(1)}(T)$ let $b_i^\star(T)=-1$ if $w_i^{(-1)}(T)> w_i^{(1)}(T)$, and $b_i^\star(T)=1$ otherwise. Now, using the list of optimal parent states, $\{z_i^{(-1),\text{parent}}(t)\}_{t=1}^{T-1}$ and $\{z_i^{(1),\text{parent}}(t)\}_{t=1}^{T-1}$, the optimal wealth-maximizing trading trajectory can be obtained by backpropagation, resulting in $\{b_i^\star(t)\}_{t=0}^T$. In the following, $\mathcal{F}^\text{trade}(\{ \tilde{x}_i(t) \}_{t=0}^T,\epsilon_i)$ shall abbreviate the mapping from $\{ \tilde{x}_i(t) \}_{t=0}^T$ to $\{b_i^\star(t)\}_{t=0}^T$ as a function of transaction cost level $\epsilon_i>0$. 

So far, it was discussed how to transform time series data $\{x_i(t)\}_{t=0}^T$ to $\{\tilde{x}_i(t)\}_{t=0}^T$, before computing binary $\{b_i^\star(t)\}_{t=0}^T$ as a function of scalars $\epsilon_i\geq 0$. Note that all of these steps can be parallelized channel-wise for all $i=1,\dots,n_x$. It remains to discuss (i) how to reach consensus over all  $n_x$ channels in order to produce final segmentation instants, $\{\tau_k \}_{k=1}^{\hat{K}}$, and (ii) how to appropriately select hyperparameters $\epsilon_i$.  

Here, it is proposed to reach consensus from a weighted average defined as follows,
\begin{equation}
q^\star(t)=\sum_{i=1}^{n_x} \eta_i p_i b_i^\star(t),~\forall t=0,\dots,T,\label{eq_def_bstart}
\end{equation}
with weights $\eta_i\in[0,1]$ such that $\sum_{i=1}^{n_x}\eta_i=1$, $p_1=1$ and
\begin{equation}
p_i = \text{arg} \max_{p_i\in\{1,-1\}} \sum_{t=0}^T \left| b_1^{\star}(t) + p_i b_i^{\star}(t)\right|,~\forall i=2,\dots,n_x.\label{eq_def_qi}
\end{equation}
Several comments are made. First, as a special case and treated as the default below, uniform weighting implies $\eta_i=\frac{1}{n_x}$. Alternatively, non-uniform weights may be used for trading-off importance of different sensor channels. (For a financial analogy, consider stock indices that weight different components according to their market capitalizations.)

Second, the introduction of binary $p_i\in\{-1,1\}$ is motivated by a symmetry argument. Suppose two channels with time series that are \emph{mirrored} with respect to the time-axis, i.e., $x_2(t)=-x_1(t)$. Then, $b_2(t)=-b_1(t)$ follows. Consequently, for $q^\star(t)=\sum_{i=1}^{n_x} \eta_i b_i^\star(t)$ with $\eta_i=\frac{1}{n_x}$ the result $q^\star(t)=0$ is obtained. However, this clearly is undesired. This can be seen by considering, for example, a sawtooth function for $x_1(t)$, and $x_2(t)=-x_1(t)$. Then, no changepoints at all would be identified eventhough these clearly exist. Therefore, the solution in \eqref{eq_def_bstart} is proposed, which resolves the symmetry issue. The reference prescription $p_i=1$ for $i=1$ is arbitrary. Any other $i\in\{2,\dots,n_x\}$ would also serve. Note that the introduction of $p_i$ only affects consensus finding over all channels. The optimal trading trajectories for individual channels are still determined independently from each other. As a detail, by defining binary $b_i(t)\in\{-1,1\}$, rather than the more conventional $b_i(t)\in\{0,1\}$, \eqref{eq_def_qi} can be evaluated by simple multiplication. By construction $q^\star(t)\in[-1,1]$. Then, $\{\tau_k \}_{k=1}^{\hat{K}}$ is computed as the ordered list of time instants at which $q^\star(t)$ crosses the threshold-level $0$ either from below or above. For multivariate time series segmentation this differentation is sufficient. Since $z_i(0)$ is initialized in a surrogate cash position and for a larger number of positive than negative $p_i$, the time instants listed in $\{\tau_k \}_{k=1}^{\hat{K}}$ alternatingly indicate the start of an up-trending (weightedly averaged over all channels) and a down-trending multivariate time series segment, respectively. In the following, $\mathcal{F}^\text{consensus}(\{\{ b_i^{\star}(t) \}_{t=0}^T\}_{i=1}^{n_x})$ shall abbreviate the consensus mapping from $\{\{ b_i^{\star}(t) \}_{t=0}^T\}_{i=1}^{n_x}$ to $\{\tau_k \}_{k=1}^{\hat{K}}$.

For any given time series data the selection of a suitable transaction cost level, $\epsilon_i$, is a priori not obvious. Therefore, an iterative approach is proposed. In the first iteration it is set $\epsilon_i= 0$, which is typically appropriate for data with high signal-to-noise ratio. In the second iteration it is set $\epsilon_i = \epsilon^\text{min}$, where $\epsilon^\text{min}>0$ is a hyperparameter. For all remaining iterations it is set $\epsilon_i^\text{new} = \gamma^\text{mult}\epsilon_i$, where multiplier $\gamma^\text{mult}>1$ is a hyperparameter and $\epsilon_i^\text{new}$ denotes the transaction cost level for the next iteration. Denoting iteration numbers by $j\geq 0$, the mapping to determine $\epsilon_i$ shall be abbreviated by $\mathcal{F}^\text{epsilon}(\epsilon_i,\epsilon^\text{min},\epsilon^\text{max},\gamma^\text{mult},j)$ in the following.

The result $\{b_i^\star(t)\}_{t=0}^T$ clearly is a function of $\epsilon_i$. Let the number of switches, i.e., the number of changes where $b_i^\star(t+1)\neq b_i^\star(t)$, be denoted by $\mathcal{L}(\epsilon_i)$. This is bounded by $0\leq \mathcal{L}(\epsilon_i)\leq T$. Furthermore, $\mathcal{L}(\epsilon_i)$ clearly is monotonously decreasing for increasing $\epsilon_i$. This follows from the state transitions according to \eqref{eq_def_zitp1_a}-\eqref{eq_def_zitp1_b} and the wealth-maximizing method to construct $\{b_i^\star(t)\}_{t=0}^T$. This monotonicity can be exploited for defining a simple stopping criterion. If $0 < \mathcal{L}(\epsilon_i)\leq K^\text{max}$, then the channel-wise $\epsilon_i$-iteration is stopped. Here, $K^\text{max}>0$ defines a user-defined maximum permissible number of segments. In addition, the pathological case must be handled where $\mathcal{L}(\epsilon_i)$ drops to zero during $\epsilon_i$-iterations. Then, the last $\epsilon_i$-solution yielding $\mathcal{L}(\epsilon_i)>0$ is returned. By construction this always implies $\mathcal{L}(\epsilon_i)>K^\text{max}$. Therefore, for such cases an additional merging-routine is employed, which is, however, introduced further below after the discussion of also handling the \emph{reversed} time series data. In the following, the binary mapping for terminating $\epsilon_i$-iterations shall be abbreviated by $\mathcal{F}^\text{terminate}(\{ b_i^{\star}(t) \}_{t=0}^T, K^\text{max})$.  

For the trading-inspired methodology the resulting trajectory, $\{ b_i^{\star}(t) \}_{t=0}^T$, clearly is a function of the ordering of time series data. Therefore, to generalize the \emph{reversed} time series data is also treated, before results for both directions are merged. Regarding reversed time series data treatment, the same $\epsilon_i$ is employed that was found iteratively for the original (non-reversed) time series. This selection is done for computational efficiency and justified since absolute-valued incremental changes are invariant with respect to data ordering. Resulting breakpoints shall be denoted by $\{ \tau_k^\text{flip} \}_{k=1}^{\hat{K}^\text{flip}}$. Regarding final merging of $\{ \tau_k \}_{k=1}^{\hat{K}}$ and $\{ \tau_k^\text{flip} \}_{k=1}^{\hat{K}^\text{flip}}$, three comments are made. First, in case two breakpoints are closer than a small hyperparameter, $\gamma^\text{close}>0$, the two breakpoints are averaged. For perspective on the order of magnitude, in numerical experiments it was selected $\gamma^\text{close}=\max(0.01T,2)$. Second, for an efficient implementation of the merging, one can exploit the fact that both lists $\{ \tau_k \}_{k=1}^{\hat{K}}$ and $\{ \tau_k^\text{flip} \}_{k=1}^{\hat{K}^\text{flip}}$ are already ordered and increasing. Furthermore, for both it typically holds that $\hat{K}\ll T$  and $\hat{K}^\text{flip}\ll T$. For perspective, for the last experiment reported in this paper it is $\hat{K}=9$ and $T=2709$. Third, after above method of merging, the number of breakpoints may still exceed a desired upper bound $K^\text{max}>0$. Then, closest breakpoints with respect to their predecessor are iteratively removed until the condition $\hat{K}\leq K^\text{max}$ is satisfied for the merged list. In the following, the full mapping implementing both merging and removal of close breakpoints shall be abbreviated by $\mathcal{F}^\text{merge}( \{ \tau_k \}_{k=1}^{\hat{K}}, \{ \tau_k^\text{flip} \}_{k=1}^{\hat{K}^\text{flip}}, \gamma^\text{close}, K^\text{max} )$.

Algorthim \ref{alg_batchOffline} summarizes above discussion. It is called APTS (A Posteriori Trading-inspired Segmentation).\\[-1pt]

\vspace{-0.2cm}
\begin{algorithm}
\SetKwInOut{Subfunctions}{\textbf{Subfunctions}}
\SetKwInOut{Input}{\textbf{Data Input}}
\SetKwInOut{Hyperparameters}{\textbf{Hyperparam.\hspace{0.05cm}}}
\SetKwInOut{Output}{\textbf{Final Result}}
\DontPrintSemicolon
\vspace{0.15cm}
\Subfunctions{$\mathcal{F}^\text{normalize}(\cdot)$, $\mathcal{F}^\text{trade}(\cdot)$, $\mathcal{F}^\text{consensus}(\cdot)$, $\mathcal{F}^\text{epsilon}(\cdot)$, $\mathcal{F}^\text{terminate}(\cdot)$, $\mathcal{F}^\text{merge}(\cdot)$.}
\vspace{0.15cm}\hrule\vspace{0.15cm}
\Hyperparameters{$\epsilon^\text{min}$, $\epsilon^\text{max}$, $\gamma^\text{mult}$, $\gamma^\text{close}$, $K^\text{max}$.}
\vspace{0.15cm}\hrule\vspace{0.15cm}
\Input{$\{\{x_i(t)\}_{t=0}^T\}_{i=0}^{n_x}$.}
\vspace{0.15cm}\hrule\vspace{0.15cm}
\For{$i\in\{1,\dots,n_x\}$}
{
$\{ \tilde{x}_i(t) \}_{t=0}^T \leftarrow \mathcal{F}^\text{normalize}\left( \{ x_i(t) \}_{t=0}^T \right)$.\;
$\epsilon_i\leftarrow 0$, $j\leftarrow 0$.\;
\While{continue}
{
$\epsilon_i \leftarrow \mathcal{F}^\text{epsilon}(\epsilon_i,\epsilon^\text{min},\epsilon^\text{max},\gamma^\text{mult},j)$.\;
$\{ b_i^{\star}(t) \}_{t=0}^T \leftarrow \mathcal{F}^\text{trade}\left( \{\tilde{x}_i(t)\}_{t=0}^T,\epsilon_i\right)$.\;
\If{$\mathcal{F}^\text{terminate}(\{ b_i^{\star}(t) \}_{t=0}^T, K^\text{max})==\text{True}$}
{
break.\;
}
$j\leftarrow j+1$.\;
} 
} 
$\{ \tau_k \}_{k=1}^{\hat{K}} \leftarrow \mathcal{F}^\text{consensus}( \{\{ b_i^{\star}(t) \}_{t=0}^T\}_{i=1}^{n_x} ).$\;
\vspace{0.15cm}\hrule\vspace{0.15cm} 
$\{ \tilde{x}_i^\text{flip}(t) \}_{t=0}^T \leftarrow \mathcal{F}^\text{normalize}\left( \{ x_i(t) \}_{t=T}^0 \right)$.\;
\For{$i\in\{1,\dots,n_x\}$}
{
$\{ b_i^{\star,\text{flip}}(t) \}_{t=T}^0 \leftarrow \mathcal{F}^\text{trade}\left( \{\tilde{x}_i^\text{flip}(t)\}_{t=0}^T,\epsilon_i\right)$.\;
} 
$\{ \tau_k^\text{flip} \}_{k=1}^{\hat{K}^\text{flip}} \leftarrow \mathcal{F}^\text{consensus}( \{\{ b_i^{\star,\text{flip}}(t) \}_{t=0}^T\}_{i=1}^{n_x} ).$\;
\vspace{0.15cm}\hrule\vspace{0.15cm} 
$\{ \tau_k \}_{k=1}^{\hat{K}} \leftarrow \mathcal{F}^\text{merge}( \{ \tau_k \}_{k=1}^{\hat{K}}, \{ \tau_k^\text{flip} \}_{k=1}^{\hat{K}^\text{flip}}, \gamma^\text{close}, K^\text{max} ).$\;
\vspace{0.15cm}\hrule\vspace{0.15cm}
\Output{$\{ \tau_k \}_{k=1}^{\hat{K}}$.}
%
%
\caption{APTS}\label{alg_batchOffline}
\end{algorithm}

Three additional comments are made. First, the dominating complexity of Algorithm \ref{alg_batchOffline} is linear in time series length and dimension, i.e., $\mathcal{O}\left( n_xT N_\epsilon\right)$. The maximum number of $\epsilon_i$-iterations, $N_\epsilon>0$, in Steps 3-9 can be determined analytically from the formula, $\epsilon^\text{min}{\gamma^\text{mult}}^{N_\epsilon-1}<\epsilon^\text{max}$. 

Second, the necessity of a normalization method resulting in channel-wise positive time series data is motivated by counterexample. As part of \eqref{eq_def_zitp1_a}, the wealth equation $w_i(t+1)=n_i(t+1)\tilde{x}_i(t+1)=\frac{c_i(t)(1-\epsilon_i)}{\tilde{x}_i(t)}\tilde{x}_i(t+1)$ occurs. Suppose $\tilde{x}_i(t)<0$ and futher $\tilde{x}_i(t+1)<\tilde{x}_i(t)$. Then, a drop in ``stock'' price from time $t$ to $t+1$ would imply an increase in wealth (due to the ratio of two negative numbers in $w_i(t+1)$). This scenario would be undesired since misleading. It is avoidable by a normalization method rendering data positive. (This excludes the standard z-normalization technique \cite{goldin1995similarity}.) 

Finally, an extension of APTS is discussed. By design Algorithm \ref{alg_batchOffline} is particularly suitable to segment time series with different local maxima and minima. However, by adding a simple data pre- and post-processing step also time series exhibiting channel-wise quasi-constant values for prolonged periods of times (plateaus) can be handled efficiently. Therefore, a small hyperparameter $\gamma^\text{plat}\geq 0$ is introduced (e.g., $\gamma^\text{plat}=0.05$) that is used after normalization in Step 2 for downsampling. Thus, a reduced time series is produced, $\{ \bar{x}_i(t) \}_{t=0}^{\bar{T}} \in \{ \{ \tilde{x}_i(t) \}_{t=0}^{T} : \tilde{x}_i(t)-\tilde{x}_i(t-1)>\gamma^\text{plat},\forall t=0,\dots,T \}$. The same filtering step is also applied analogously after Step 11 to the reversed time series. The effect is removal of plateau-like data parts. Then, after processing via $\mathcal{F}^\text{trade}(\{\bar{x}_i(t)\}_{t=0}^{\bar{T}},\epsilon_i)$, the previously removed time periods must be reinserted to recover the original time indexing and produce an adjusted $\{ b_i^\star(t)\}_{t=0}^T$ in Step 6, and analogously in Step 13 for the reversed time series. This methodology permits to accurately identify start and end times of plateau-like segments in time series data.

\subsection{Two Comparative Algorithms from the Literature\label{subsec_2CompMethods}}

Before reporting numerical results, two comparative algorithms from the literature are briefly discussed.

Based on the fundamental belief that underlying data in each segment can be well approximated by piecewise affine representations, according to \cite{keogh2001online} most time series segmentation algorithms can be grouped into one of three categories: \emph{sliding window}-approaches (a segment is grown until exceeding some error bound and starting a new segment), \emph{top-down}-approaches (a time series is recursively partitioned until a stopping criterion is met), and \emph{bottom-up}-approaches (starting from the finest possible approximation, segments are merged until a stopping criterion is met). In the same reference it is found empirically that the bottom-up approach (BU) is most suitable. Therefore, BU according to \cite{keogh2001online} is here used for comparison. For BU and for the fitting of piecewise affine representations, local solutions of either linear interpolation or linear regression (least-squares) problems are required for each segment. In \cite{keogh2001online}, the latter approach is used for final experiments. Here, for a Python implementation numpy's \texttt{linalg.lstsq}-solver is employed for the solution of least-squares problems. It is worthwhile noting that the original reference exclusively focused on one-dimensional time series data (i.e., $n_x=1$) for empirical evaluation. In this paper, BU is altered for handling of multivariate time series by straightforward extension. As stopping criterion, a desired number of segments is prescribed. This number represents the only hyperparameter. Thus, as soon as sufficiently many segments are merged (starting from the finest possible approximation) BU stops. 

In \cite{hallac2019greedy}, a top-down algorithm labeled \emph{Greedy Gaussian Segmentation} (GGS) is proposed. Fundamental belief is that underlying data in each segment can be explained as samples from a multivariate Gaussian distribution with constant mean and covariance for each segment. To determine breakpoints (i.e., segment boundaries) a covariance-regularized maximum likelihood objective function is formulated,
\begin{align}
&\max_{\tau_1,\dots,\tau_K}-\frac{1}{2} \sum_{k=1}^{K+1} (\tau_k-\tau_{k-1})\text{log}\text{det} (S^{(k)} + \frac{\lambda I}{\tau_k-\tau_{k-1}}) -\notag\\
&\hspace*{2cm} \dots - \lambda\text{Tr}(S^{(k)} + \frac{\lambda I}{\tau_k-\tau_{k-1}})^{-1} , \label{eq_GGS}
\end{align}
with hyperparameter $K$ indicating the number of breakpoints, regularization hyperparameter $\lambda>0$, trace-operator $\text{Tr}(\cdot)$, empirical covariance $S^{(k)}=\frac{1}{\tau_k-\tau_{k-1}}\sum_{t=\tau_{k-1}}^{\tau_k-1} (x(t)-\mu^{(i)})(x(t)-\mu^{(i)})^T$, empirical mean $\mu^{(i)}=\frac{1}{\tau_k-\tau_{k-1}}\sum_{t=\tau_{k-1}}^{\tau_k-1} x(t)$, data $x(t)\in\mathbb{R}^{n_x}$ and $I$ denoting the identity matrix. By alternatingly adding new breakpoints and adjusting the position of all breakpoints \eqref{eq_GGS} is maximized until no change of any breakpoint further improves the objective function that sums costs over all segments. GGS requires selection of two hyperparameters. These are the desired number of breakpoints, $K>0$, and the regularization parameter, $\lambda>0$, to enforce positive definite (invertible) estimated covariance matrices. Code for GGS is publicly available and taken from the authors' website to conduct below numerical experiments. Non-trivial library routines required for the implementation of GGS include numpy's \texttt{linalg.slogdet}, \texttt{linalg.inv} and \texttt{linalg.cholesky}-functions. For a fixed number of breakpoints, $K>0$, the complexity is $\mathcal{O}(KLn_x^3 T)$ flops, where $L$ is the average number of iterations required for adjusting breakpoints. While an upper bound on $L$ is not known, it was observed empirically that it is modest when $K$ is not too large \cite{hallac2019greedy}. The cubic complexity in the dimension of the multivariate time series $n_x$ is underlined.

\section{Numerical Results\label{sec_expts}}

\subsection{Selection of Experiments and Evaluation of Results}

Experimental data was selected with the purpose of comparing BU \cite{keogh2001online}, GGS \cite{hallac2019greedy}, and APTS on (i) a variety of different shapes of time series data to analyze generalizability of these methods, and (ii) on a large-scale example to compare computational efficiencies. Therefore, in addition to two synthetic examples eight samples from the publicly accessible UCR time series archive \cite{dau2019ucr} were selected. This combination enabled testing on time series data with piecewise-affine segments, half-circles, rising and falling segments with local maxima and minima, plateau-like segments, strongly fluctuating data with different noise levels, impulse-like data, noise-perturbed  multivariate data, and large-scale data with $n_x=1000$ channels and length $T=2709$.

Two methods for evaluation of results were considered: (i) manually determining ``ground-truth'' segmentation instants before evaluating an error metric, and (ii) a visual approach where all results are explicitly plotted for subjective inspection and evaluation by the reader. The first approach has several disadvantages. These include that determining a ground-truth is subjective to the selector's preference. Furthermore, summarizing performance by a single scalar error metric may insufficiently capture actual segmentation quality of algorithms. Therefore, the second visual approach is taken, which is also appropriate given the typical unsupervised nature of time series segmentation tasks. In addition, computational solve times are reported for each experiment.

\subsection{Selection of Hyperparameters}

In order to demonstrate robustness of APTS and adaptability to different shapes of data a single set of hyperparameters is used throughout all experiments. This set is summarized in Table \ref{tab_hyperparam}. For emphasis of handling of data with plateau-like segments, results for Examples 4-6 are also reported for $\gamma^\text{plat}=0.05$, besides the default solutions for $\gamma^\text{plat}=0$.

\begin{table}
\renewcommand{\arraystretch}{1.2}
\vspace{0.6cm}
\centering
\begin{tabular}{|l|l|}
\hline
 Symbol & Value \\ 
\hline
$\epsilon^\text{min}$ & 0.01 \\ 
$\epsilon^\text{max}$ & 1 \\ 
$\gamma^\text{mult}$ & 2 \\ 
$\gamma^\text{close}$ & $\max(0.01T,2)$ \\ 
$\gamma^\text{plat}$ & $0$ \\ 
$K^\text{max}$ & 10 \\      
\hline
\end{tabular}
\caption{To demonstrate robustness of APTS and adaptability to various time series data a single set of hyperparameters is used throughout all Experiments 1-10. For special identification of plateau-like segments results for Experiments 4-6 are additionally reported for $\gamma^\text{plat}=0.05$.}
\label{tab_hyperparam}
\end{table}

APTS does not define a specific number of segments by a hyperparameter selection. Instead, an upper bound $K^\text{max}>0$ must be provided, before Algorithm \ref{alg_batchOffline} returns $\hat{K}\leq K^\text{max}$. In contrast, without adding on-top iterative selection methods BU and GGS require prescription of a desired number of segments $K>0$. Therefore, in order to compare BU, GGS and APTS the following method is applied. First, APTS is run. Then, BU and GGS are run with setting $K=\hat{K}$. This permits a comparison for the same number of segments.

BU does not require a hyperparameter selection beyond $K$. In contrast, GGS additionally requires setting of the regularization hyperparameter $\lambda>0$ in \eqref{eq_GGS}. After empirical testing, $\lambda=10^{-1}$ was determined for Examples 1-9 and $\lambda=10^{-4}$ for Example 10. For the latter experiment $\lambda$ had to be reduced in order to produce solutions with larger $K$. (The final result is obtained for $K=9$.)

\subsection{Numerical Results}

\begin{figure}
\centering
\vspace{0.3cm}
\begin{tikzpicture}

\begin{groupplot}[group style={group size=1 by 3,vertical sep=0.55cm}]

\nextgroupplot[
title style={yshift=-0.25cm,},
title={\hspace{-6.01cm}BU},
xticklabels={\empty},
ylabel={\small{$x(t)$}},
xmin=0, xmax=99,
ymin=-3.4, ymax=26.2969696969697,
width=8.2cm,
height=3cm,
tick align=outside,
tick pos=left,
x grid style={lightgray!92.02614379084967!black},
y grid style={lightgray!92.02614379084967!black}
]
\addplot [blue, forget plot]
table {%
0 0
1 0.806060606060606
2 1.61212121212121
3 2.41818181818182
4 3.22424242424242
5 4.03030303030303
6 4.83636363636364
7 5.64242424242424
8 6.44848484848485
9 7.25454545454545
10 8.06060606060606
11 8.86666666666667
12 9.67272727272727
13 10.4787878787879
14 11.2848484848485
15 12.0909090909091
16 12.8969696969697
17 12.7939393939394
18 11.7818181818182
19 10.769696969697
20 9.75757575757576
21 8.74545454545455
22 7.73333333333333
23 6.72121212121212
24 5.70909090909091
25 4.6969696969697
26 3.68484848484848
27 2.67272727272727
28 1.66060606060606
29 0.648484848484848
30 -0.363636363636363
31 -1.37575757575757
32 -2.38787878787879
33 -3.4
34 -1.98787878787879
35 -0.575757575757576
36 0.836363636363637
37 2.24848484848485
38 3.66060606060606
39 5.07272727272727
40 6.48484848484848
41 7.8969696969697
42 9.30909090909091
43 10.7212121212121
44 12.1333333333333
45 13.5454545454545
46 14.9575757575758
47 16.369696969697
48 17.7818181818182
49 19.1939393939394
50 19.3939393939394
51 18.3818181818182
52 17.369696969697
53 16.3575757575758
54 15.3454545454545
55 14.3333333333333
56 13.3212121212121
57 12.3090909090909
58 11.2969696969697
59 10.2848484848485
60 9.27272727272727
61 8.26060606060606
62 7.24848484848485
63 6.23636363636364
64 5.22424242424243
65 4.21212121212121
66 3.2
67 4.61212121212121
68 6.02424242424243
69 7.43636363636364
70 8.84848484848485
71 10.2606060606061
72 11.6727272727273
73 13.0848484848485
74 14.4969696969697
75 15.9090909090909
76 17.3212121212121
77 18.7333333333333
78 20.1454545454545
79 21.5575757575758
80 22.969696969697
81 24.3818181818182
82 25.7939393939394
83 26.2969696969697
84 25.8909090909091
85 25.4848484848485
86 25.0787878787879
87 24.6727272727273
88 24.2666666666667
89 23.8606060606061
90 23.4545454545455
91 23.0484848484848
92 22.6424242424242
93 22.2363636363636
94 21.830303030303
95 21.4242424242424
96 21.0181818181818
97 20.6121212121212
98 20.2060606060606
99 19.8
};
\addplot [red, dashed, forget plot]
table {%
17 -3.4
17 26.2969696969697
};
\addplot [red, dashed, forget plot]
table {%
33 -3.4
33 26.2969696969697
};
\addplot [red, dashed, forget plot]
table {%
50 -3.4
50 26.2969696969697
};
\addplot [red, dashed, forget plot]
table {%
66 -3.4
66 26.2969696969697
};
\addplot [red, dashed, forget plot]
table {%
83 -3.4
83 26.2969696969697
};
\nextgroupplot[
title style={yshift=-0.25cm,},
title={\hspace{-5.82cm}GGS},
xticklabels={\empty},
ylabel={\small{$x(t)$}},
xmin=0, xmax=99,
ymin=-3.4, ymax=26.2969696969697,
width=8.2cm,
height=3cm,
tick align=outside,
tick pos=left,
x grid style={lightgray!92.02614379084967!black},
y grid style={lightgray!92.02614379084967!black}
]
\addplot [blue,  forget plot]
table {%
0 0
1 0.806060606060606
2 1.61212121212121
3 2.41818181818182
4 3.22424242424242
5 4.03030303030303
6 4.83636363636364
7 5.64242424242424
8 6.44848484848485
9 7.25454545454545
10 8.06060606060606
11 8.86666666666667
12 9.67272727272727
13 10.4787878787879
14 11.2848484848485
15 12.0909090909091
16 12.8969696969697
17 12.7939393939394
18 11.7818181818182
19 10.769696969697
20 9.75757575757576
21 8.74545454545455
22 7.73333333333333
23 6.72121212121212
24 5.70909090909091
25 4.6969696969697
26 3.68484848484848
27 2.67272727272727
28 1.66060606060606
29 0.648484848484848
30 -0.363636363636363
31 -1.37575757575757
32 -2.38787878787879
33 -3.4
34 -1.98787878787879
35 -0.575757575757576
36 0.836363636363637
37 2.24848484848485
38 3.66060606060606
39 5.07272727272727
40 6.48484848484848
41 7.8969696969697
42 9.30909090909091
43 10.7212121212121
44 12.1333333333333
45 13.5454545454545
46 14.9575757575758
47 16.369696969697
48 17.7818181818182
49 19.1939393939394
50 19.3939393939394
51 18.3818181818182
52 17.369696969697
53 16.3575757575758
54 15.3454545454545
55 14.3333333333333
56 13.3212121212121
57 12.3090909090909
58 11.2969696969697
59 10.2848484848485
60 9.27272727272727
61 8.26060606060606
62 7.24848484848485
63 6.23636363636364
64 5.22424242424243
65 4.21212121212121
66 3.2
67 4.61212121212121
68 6.02424242424243
69 7.43636363636364
70 8.84848484848485
71 10.2606060606061
72 11.6727272727273
73 13.0848484848485
74 14.4969696969697
75 15.9090909090909
76 17.3212121212121
77 18.7333333333333
78 20.1454545454545
79 21.5575757575758
80 22.969696969697
81 24.3818181818182
82 25.7939393939394
83 26.2969696969697
84 25.8909090909091
85 25.4848484848485
86 25.0787878787879
87 24.6727272727273
88 24.2666666666667
89 23.8606060606061
90 23.4545454545455
91 23.0484848484848
92 22.6424242424242
93 22.2363636363636
94 21.830303030303
95 21.4242424242424
96 21.0181818181818
97 20.6121212121212
98 20.2060606060606
99 19.8
};
\addplot [red, dashed, forget plot]
table {%
8 -3.4
8 26.2969696969697
};
\addplot [red, dashed, forget plot]
table {%
25 -3.4
25 26.2969696969697
};
\addplot [red, dashed, forget plot]
table {%
40 -3.4
40 26.2969696969697
};
\addplot [red, dashed, forget plot]
table {%
79 -3.4
79 26.2969696969697
};
\addplot [red, dashed, forget plot]
table {%
93 -3.4
93 26.2969696969697
};
\nextgroupplot[
title style={yshift=-0.25cm,},
title={\hspace{-5.68cm}APTS},
xlabel={\small{$t$}},
ylabel={\small{$x(t)$}},
xmin=0, xmax=99,
ymin=-3.4, ymax=26.2969696969697,
width=8.2cm,
height=3cm,
tick align=outside,
tick pos=left,
x grid style={lightgray!92.02614379084967!black},
y grid style={lightgray!92.02614379084967!black}
]
\addplot [blue,   forget plot]
table {%
0 0
1 0.806060606060606
2 1.61212121212121
3 2.41818181818182
4 3.22424242424242
5 4.03030303030303
6 4.83636363636364
7 5.64242424242424
8 6.44848484848485
9 7.25454545454545
10 8.06060606060606
11 8.86666666666667
12 9.67272727272727
13 10.4787878787879
14 11.2848484848485
15 12.0909090909091
16 12.8969696969697
17 12.7939393939394
18 11.7818181818182
19 10.769696969697
20 9.75757575757576
21 8.74545454545455
22 7.73333333333333
23 6.72121212121212
24 5.70909090909091
25 4.6969696969697
26 3.68484848484848
27 2.67272727272727
28 1.66060606060606
29 0.648484848484848
30 -0.363636363636363
31 -1.37575757575757
32 -2.38787878787879
33 -3.4
34 -1.98787878787879
35 -0.575757575757576
36 0.836363636363637
37 2.24848484848485
38 3.66060606060606
39 5.07272727272727
40 6.48484848484848
41 7.8969696969697
42 9.30909090909091
43 10.7212121212121
44 12.1333333333333
45 13.5454545454545
46 14.9575757575758
47 16.369696969697
48 17.7818181818182
49 19.1939393939394
50 19.3939393939394
51 18.3818181818182
52 17.369696969697
53 16.3575757575758
54 15.3454545454545
55 14.3333333333333
56 13.3212121212121
57 12.3090909090909
58 11.2969696969697
59 10.2848484848485
60 9.27272727272727
61 8.26060606060606
62 7.24848484848485
63 6.23636363636364
64 5.22424242424243
65 4.21212121212121
66 3.2
67 4.61212121212121
68 6.02424242424243
69 7.43636363636364
70 8.84848484848485
71 10.2606060606061
72 11.6727272727273
73 13.0848484848485
74 14.4969696969697
75 15.9090909090909
76 17.3212121212121
77 18.7333333333333
78 20.1454545454545
79 21.5575757575758
80 22.969696969697
81 24.3818181818182
82 25.7939393939394
83 26.2969696969697
84 25.8909090909091
85 25.4848484848485
86 25.0787878787879
87 24.6727272727273
88 24.2666666666667
89 23.8606060606061
90 23.4545454545455
91 23.0484848484848
92 22.6424242424242
93 22.2363636363636
94 21.830303030303
95 21.4242424242424
96 21.0181818181818
97 20.6121212121212
98 20.2060606060606
99 19.8
};
\addplot [red, dashed, forget plot]
table {%
16 -3.4
16 26.2969696969697
};
\addplot [red, dashed, forget plot]
table {%
33 -3.4
33 26.2969696969697
};
\addplot [red, dashed, forget plot]
table {%
50 -3.4
50 26.2969696969697
};
\addplot [red, dashed, forget plot]
table {%
66 -3.4
66 26.2969696969697
};
\addplot [red, dashed, forget plot]
table {%
83 -3.4
83 26.2969696969697
};
\end{groupplot}

\end{tikzpicture}
\caption{Example 1. Results for synthetic one-dimensional time series data composed of piecewise affine segments. Vertical red dashed lines indicate identified segmentation indices.}
\label{fig_Ex1}
\end{figure}
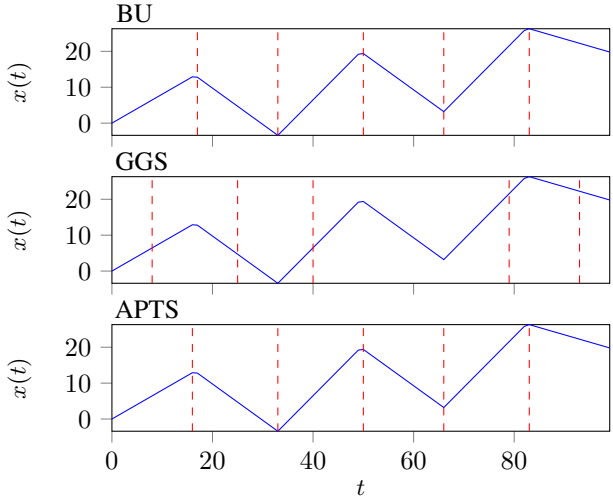

\begin{figure}
\centering
\vspace{0.3cm}
\begin{tikzpicture}

\begin{groupplot}[group style={group size=1 by 3,vertical sep=0.55cm}]

\nextgroupplot[
title style={yshift=-0.25cm,},
title={\hspace{-6.01cm}BU},
xticklabels={\empty},
ylabel={\small{$x(t)$}},
xmin=0, xmax=99,
ymin=-0.6, ymax=16.7,
width=8.2cm,
height=3cm,
tick align=outside,
tick pos=left,
x grid style={lightgray!92.02614379084967!black},
y grid style={lightgray!92.02614379084967!black}
]
\addplot [blue, forget plot]
table {%
0 2.02066721859313e-15
1 5.65651354944167
2 7.87269125042682
3 9.4864546125897
4 10.769602246926
5 11.8316296906899
6 12.7271093081644
7 13.4903381690583
8 14.1421134431402
9 14.6968409445308
10 15.1653126888874
11 15.555684895414
12 15.8741341011718
13 16.124037143421
14 16.3089883593023
15 16.4312493142005
16 16.4919346305075
17 16.4919346305075
18 16.4312493142005
19 16.3089883593023
20 16.124037143421
21 15.8741341011718
22 15.555684895414
23 15.1653126888874
24 14.6968409445308
25 14.1421134431402
26 13.4903381690583
27 12.7271093081644
28 11.8316296906899
29 10.769602246926
30 9.4864546125897
31 7.87269125042682
32 5.65651354944167
33 1.99840144432528e-15
34 5.65651354944167
35 7.87269125042682
36 9.4864546125897
37 10.769602246926
38 11.8316296906899
39 12.7271093081644
40 13.4903381690583
41 14.1421134431402
42 14.6968409445308
43 15.1653126888874
44 15.555684895414
45 15.8741341011718
46 16.124037143421
47 16.3089883593023
48 16.4312493142005
49 16.4919346305075
50 16.4919346305075
51 16.4312493142005
52 16.3089883593023
53 16.124037143421
54 15.8741341011718
55 15.555684895414
56 15.1653126888874
57 14.6968409445308
58 14.1421134431402
59 13.4903381690583
60 12.7271093081644
61 11.8316296906899
62 10.769602246926
63 9.4864546125897
64 7.87269125042682
65 5.65651354944165
66 1.99840144432528e-15
67 5.65651354944166
68 7.87269125042681
69 9.4864546125897
70 10.769602246926
71 11.8316296906899
72 12.7271093081644
73 13.4903381690583
74 14.1421134431402
75 14.6968409445308
76 15.1653126888874
77 15.555684895414
78 15.8741341011718
79 16.124037143421
80 16.3089883593023
81 16.4312493142005
82 16.4919346305075
83 16.4919346305075
84 16.4312493142005
85 16.3089883593023
86 16.124037143421
87 15.8741341011718
88 15.555684895414
89 15.1653126888874
90 14.6968409445308
91 14.1421134431402
92 13.4903381690582
93 12.7271093081644
94 11.8316296906899
95 10.769602246926
96 9.4864546125897
97 7.87269125042682
98 5.65651354944165
99 0.28724285632306
};
\addplot [red, dashed, forget plot]
table {%
14 1.99840144432528e-15
14 16.4919346305075
};
\addplot [red, dashed, forget plot]
table {%
33 1.99840144432528e-15
33 16.4919346305075
};
\addplot [red, dashed, forget plot]
table {%
38 1.99840144432528e-15
38 16.4919346305075
};
\addplot [red, dashed, forget plot]
table {%
65 1.99840144432528e-15
65 16.4919346305075
};
\addplot [red, dashed, forget plot]
table {%
80 1.99840144432528e-15
80 16.4919346305075
};
\nextgroupplot[
title style={yshift=-0.25cm,},
title={\hspace{-5.82cm}GGS},
xticklabels={\empty},
ylabel={\small{$x(t)$}},
xmin=0, xmax=99,
ymin=-0.6, ymax=16.7,
width=8.2cm,
height=3cm,
tick align=outside,
tick pos=left,
x grid style={lightgray!92.02614379084967!black},
y grid style={lightgray!92.02614379084967!black}
]
\addplot [blue,  forget plot]
table {%
0 2.02066721859313e-15
1 5.65651354944167
2 7.87269125042682
3 9.4864546125897
4 10.769602246926
5 11.8316296906899
6 12.7271093081644
7 13.4903381690583
8 14.1421134431402
9 14.6968409445308
10 15.1653126888874
11 15.555684895414
12 15.8741341011718
13 16.124037143421
14 16.3089883593023
15 16.4312493142005
16 16.4919346305075
17 16.4919346305075
18 16.4312493142005
19 16.3089883593023
20 16.124037143421
21 15.8741341011718
22 15.555684895414
23 15.1653126888874
24 14.6968409445308
25 14.1421134431402
26 13.4903381690583
27 12.7271093081644
28 11.8316296906899
29 10.769602246926
30 9.4864546125897
31 7.87269125042682
32 5.65651354944167
33 1.99840144432528e-15
34 5.65651354944167
35 7.87269125042682
36 9.4864546125897
37 10.769602246926
38 11.8316296906899
39 12.7271093081644
40 13.4903381690583
41 14.1421134431402
42 14.6968409445308
43 15.1653126888874
44 15.555684895414
45 15.8741341011718
46 16.124037143421
47 16.3089883593023
48 16.4312493142005
49 16.4919346305075
50 16.4919346305075
51 16.4312493142005
52 16.3089883593023
53 16.124037143421
54 15.8741341011718
55 15.555684895414
56 15.1653126888874
57 14.6968409445308
58 14.1421134431402
59 13.4903381690583
60 12.7271093081644
61 11.8316296906899
62 10.769602246926
63 9.4864546125897
64 7.87269125042682
65 5.65651354944165
66 1.99840144432528e-15
67 5.65651354944166
68 7.87269125042681
69 9.4864546125897
70 10.769602246926
71 11.8316296906899
72 12.7271093081644
73 13.4903381690583
74 14.1421134431402
75 14.6968409445308
76 15.1653126888874
77 15.555684895414
78 15.8741341011718
79 16.124037143421
80 16.3089883593023
81 16.4312493142005
82 16.4919346305075
83 16.4919346305075
84 16.4312493142005
85 16.3089883593023
86 16.124037143421
87 15.8741341011718
88 15.555684895414
89 15.1653126888874
90 14.6968409445308
91 14.1421134431402
92 13.4903381690582
93 12.7271093081644
94 11.8316296906899
95 10.769602246926
96 9.4864546125897
97 7.87269125042682
98 5.65651354944165
99 0.28724285632306
};
\addplot [red, dashed, forget plot]
table {%
5 1.99840144432528e-15
5 16.4919346305075
};
\addplot [red, dashed, forget plot]
table {%
10 1.99840144432528e-15
10 16.4919346305075
};
\addplot [red, dashed, forget plot]
table {%
24 1.99840144432528e-15
24 16.4919346305075
};
\addplot [red, dashed, forget plot]
table {%
75 1.99840144432528e-15
75 16.4919346305075
};
\addplot [red, dashed, forget plot]
table {%
91 1.99840144432528e-15
91 16.4919346305075
};
\nextgroupplot[
title style={yshift=-0.25cm,},
title={\hspace{-5.68cm}APTS},
xlabel={\small{$t$}},
ylabel={\small{$x(t)$}},
xmin=0, xmax=99,
ymin=-0.6, ymax=16.7,
width=8.2cm,
height=3cm,
tick align=outside,
tick pos=left,
x grid style={lightgray!92.02614379084967!black},
y grid style={lightgray!92.02614379084967!black}
]
\addplot [blue, forget plot]
table {%
0 2.02066721859313e-15
1 5.65651354944167
2 7.87269125042682
3 9.4864546125897
4 10.769602246926
5 11.8316296906899
6 12.7271093081644
7 13.4903381690583
8 14.1421134431402
9 14.6968409445308
10 15.1653126888874
11 15.555684895414
12 15.8741341011718
13 16.124037143421
14 16.3089883593023
15 16.4312493142005
16 16.4919346305075
17 16.4919346305075
18 16.4312493142005
19 16.3089883593023
20 16.124037143421
21 15.8741341011718
22 15.555684895414
23 15.1653126888874
24 14.6968409445308
25 14.1421134431402
26 13.4903381690583
27 12.7271093081644
28 11.8316296906899
29 10.769602246926
30 9.4864546125897
31 7.87269125042682
32 5.65651354944167
33 1.99840144432528e-15
34 5.65651354944167
35 7.87269125042682
36 9.4864546125897
37 10.769602246926
38 11.8316296906899
39 12.7271093081644
40 13.4903381690583
41 14.1421134431402
42 14.6968409445308
43 15.1653126888874
44 15.555684895414
45 15.8741341011718
46 16.124037143421
47 16.3089883593023
48 16.4312493142005
49 16.4919346305075
50 16.4919346305075
51 16.4312493142005
52 16.3089883593023
53 16.124037143421
54 15.8741341011718
55 15.555684895414
56 15.1653126888874
57 14.6968409445308
58 14.1421134431402
59 13.4903381690583
60 12.7271093081644
61 11.8316296906899
62 10.769602246926
63 9.4864546125897
64 7.87269125042682
65 5.65651354944165
66 1.99840144432528e-15
67 5.65651354944166
68 7.87269125042681
69 9.4864546125897
70 10.769602246926
71 11.8316296906899
72 12.7271093081644
73 13.4903381690583
74 14.1421134431402
75 14.6968409445308
76 15.1653126888874
77 15.555684895414
78 15.8741341011718
79 16.124037143421
80 16.3089883593023
81 16.4312493142005
82 16.4919346305075
83 16.4919346305075
84 16.4312493142005
85 16.3089883593023
86 16.124037143421
87 15.8741341011718
88 15.555684895414
89 15.1653126888874
90 14.6968409445308
91 14.1421134431402
92 13.4903381690582
93 12.7271093081644
94 11.8316296906899
95 10.769602246926
96 9.4864546125897
97 7.87269125042682
98 5.65651354944165
99 0.28724285632306
};
\addplot [red, dashed, forget plot]
table {%
16 1.99840144432528e-15
16 16.4919346305075
};
\addplot [red, dashed, forget plot]
table {%
33 1.99840144432528e-15
33 16.4919346305075
};
\addplot [red, dashed, forget plot]
table {%
49 1.99840144432528e-15
49 16.4919346305075
};
\addplot [red, dashed, forget plot]
table {%
66 1.99840144432528e-15
66 16.4919346305075
};
\addplot [red, dashed, forget plot]
table {%
82 1.99840144432528e-15
82 16.4919346305075
};
\end{groupplot}

\end{tikzpicture}
\caption{Example 2. Results for synthetic one-dimensional time series data composed of half-circles.}
\label{fig_Ex2}
\end{figure}
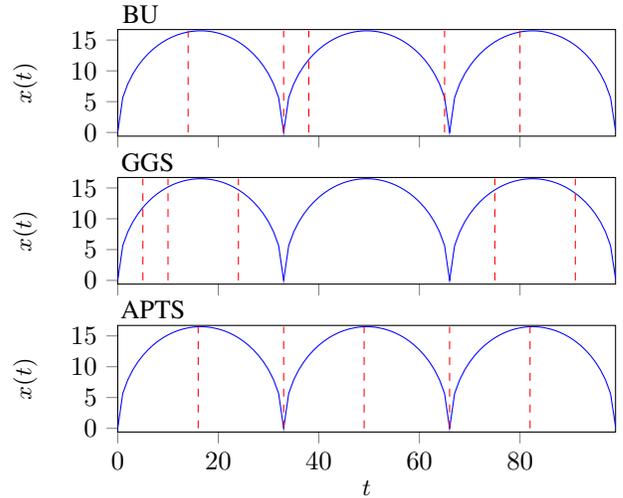

\begin{figure}
\centering
\input{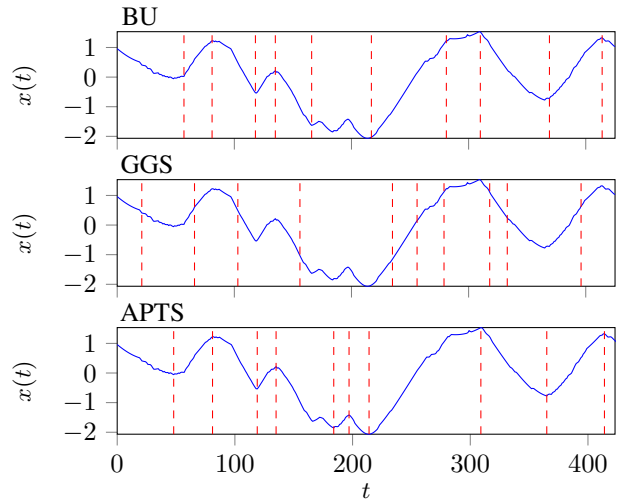}
\caption{Example 3. One-dimensional sample from dataset \textsf{'Yoga'} \cite{dau2019ucr}.}
\label{fig_Ex3}
\end{figure}

\begin{figure}
\centering
\input{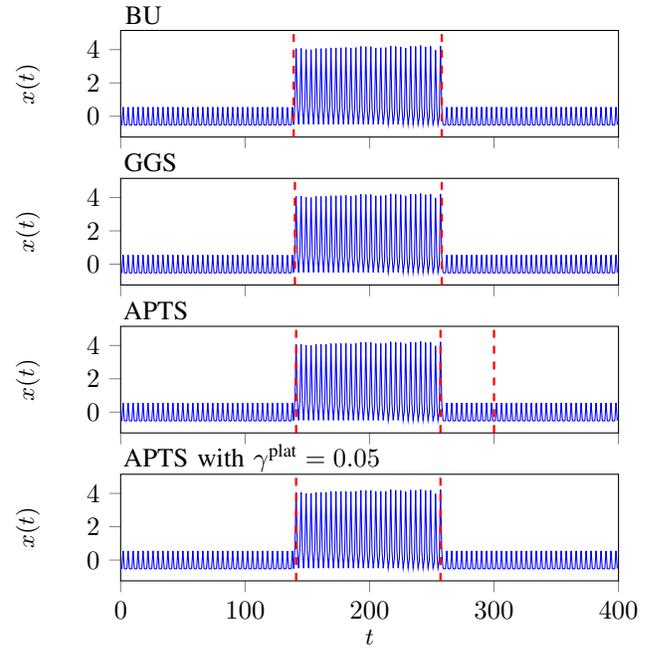}
\caption{Example 6. One-dimensional sample from dataset \textsf{'ACSF1'} \cite{dau2019ucr}. The effect of $\gamma^\text{plat}=0.05$ is further elaborated in Fig. \ref{fig_Ex6_zoom}.}
\label{fig_Ex6}
\end{figure}

\begin{figure}
\centering
\begin{tikzpicture}

\begin{axis}[
title style={yshift=-0.25cm},
title={APTS Zoom-in},
xtick={296,300,304},
xticklabels={297,300,303},
ytick={-0.532,-0.527},
yticklabels={-0.532,-0.527},
xlabel={t},
xmin=295, xmax=305,
ymin=-0.535, ymax=-0.527,
width=4cm,
height=2.5cm,
tick align=outside,
tick pos=left,
x grid style={lightgray!92.02614379084967!black},
y grid style={lightgray!92.02614379084967!black}
]
\addplot [blue, mark=*, mark size=1, mark options={solid}, forget plot]
table {%
294 0.55146978
295 -0.50822021
296 -0.53347362
297 -0.50785772
298 0.55089248
299 -0.50822021
300 -0.53393904
301 -0.50785772
302 0.55151901
303 -0.50821573
304 -0.53347362
305 -0.50785772
};
\addplot [black, dotted, forget plot]
table {%
295 -0.53347362
305 -0.53347362
};
\addplot [red, dashed, forget plot]
table {%
141 -0.53393904
141 4.2421148
};
\addplot [red, dashed, forget plot]
table {%
257 -0.53393904
257 4.2421148
};
\addplot [red, dashed, forget plot]
table {%
300 -0.53393904
300 4.2421148
};
\end{axis}

\end{tikzpicture}~
\begin{tikzpicture}

\begin{axis}[
title style={yshift=-0.25cm},
title={APTS Zoom-in},
xlabel={t},
ytick={-0.53,-0.49},
yticklabels={-0.53,-0.49},
xmin=291, xmax=309,
ymin=-0.545, ymax=-0.47,
width=4cm,
height=2.5cm,
tick align=outside,
tick pos=left,
x grid style={lightgray!92.02614379084967!black},
y grid style={lightgray!92.02614379084967!black}
]
\addplot [blue, mark=*, mark size=1, mark options={solid}, forget plot]
table {%
0 -0.53393904
1 -0.50785772
2 0.54793887
3 -0.50821573
4 -0.53393904
5 -0.50785772
6 0.54715123
7 -0.50821573
8 -0.53393904
9 -0.50785772
10 0.54824765
11 -0.50821573
12 -0.53393904
13 -0.50785772
14 0.54771511
15 -0.50821573
16 -0.53393904
17 -0.50785772
18 0.54796124
19 -0.50821573
20 -0.53347362
21 -0.50785772
22 0.54696328
23 -0.50822021
24 -0.53347362
25 -0.50785772
26 0.54739737
27 -0.50822021
28 -0.53347362
29 -0.50785772
30 0.5481134
31 -0.50822021
32 -0.53347362
33 -0.50785772
34 0.54909794
35 -0.50822021
36 -0.53393904
37 -0.50785772
38 0.54855196
39 -0.50821573
40 -0.53393904
41 -0.50785772
42 0.54757638
43 -0.50821573
44 -0.53393904
45 -0.50785772
46 0.54688272
47 -0.50821573
48 -0.53347362
49 -0.50785772
50 0.54669477
51 -0.50822021
52 -0.53254278
53 -0.50785772
54 0.52654753
55 -0.50822021
56 -0.53393904
57 -0.50785772
58 0.54873097
59 -0.50821573
60 -0.53393904
61 -0.50785772
62 0.54845351
63 -0.50821573
64 -0.53393904
65 -0.50785772
66 0.5481134
67 -0.50821573
68 -0.53393904
69 -0.50785772
70 0.54887865
71 -0.50821573
72 -0.53393904
73 -0.50785772
74 0.54800152
75 -0.50821573
76 -0.53393904
77 -0.50785772
78 0.54815367
79 -0.50821573
80 -0.53347362
81 -0.50785772
82 0.54791649
83 -0.50822021
84 -0.53347362
85 -0.50785772
86 0.54719151
87 -0.50822021
88 -0.53393904
89 -0.50785772
90 0.5476614
91 -0.50821573
92 -0.53393904
93 -0.50785772
94 0.54754953
95 -0.50821573
96 -0.53347362
97 -0.50785772
98 0.54803732
99 -0.50822021
100 -0.53347362
101 -0.50785772
102 0.54781804
103 -0.50822021
104 -0.53347362
105 -0.50785772
106 0.54775538
107 -0.50822021
108 -0.53393904
109 -0.50785772
110 0.54862804
111 -0.50821573
112 -0.53347362
113 -0.50785772
114 0.54862804
115 -0.50822021
116 -0.53347362
117 -0.50785772
118 0.54777776
119 -0.50822021
120 -0.53347362
121 -0.50785772
122 0.54739289
123 -0.50822021
124 -0.53347362
125 -0.50785772
126 0.54736157
127 -0.50822021
128 -0.53347362
129 -0.50785772
130 0.54646653
131 -0.50822021
132 -0.5330082
133 -0.50785772
134 0.53898404
135 -0.50822021
136 -0.53347362
137 -0.50785772
138 0.54791649
139 -0.50822021
140 0.71247313
141 4.0609284
142 0.51861305
143 -0.48578168
144 0.66263311
145 4.0725728
146 0.51785674
147 -0.48575931
148 0.67148501
149 4.0059644
150 0.51605772
151 -0.48605467
152 0.6481962
153 3.9938545
154 0.51634861
155 -0.4861576
156 0.59975244
157 4.063721
158 0.5174316
159 -0.48586671
160 0.5848501
161 4.0609284
162 0.51663502
163 -0.48589356
164 0.56295755
165 4.0674488
166 0.51734657
167 -0.48589356
168 0.52895518
169 4.1079715
170 0.51829083
171 -0.48574141
172 0.50100771
173 4.1070406
174 0.51846537
175 -0.48578616
176 0.42927961
177 4.1140264
178 0.51803127
179 -0.48580406
180 0.43999318
181 4.1289332
182 0.51787912
183 -0.48572351
184 0.44651351
185 4.0944654
186 0.51785227
187 -0.48590699
188 0.382702
189 4.121943
190 0.51846984
191 -0.48581301
192 0.3165634
193 4.1890169
194 0.51805813
195 -0.4855266
196 0.28395727
197 4.1769071
198 0.51837586
199 -0.485634
200 0.25834138
201 4.1792342
202 0.51834901
203 -0.48564743
204 0.28815051
205 4.1061098
206 0.51390964
207 -0.48590699
208 0.23318642
209 4.1140264
210 0.51266107
211 -0.48588909
212 0.21362543
213 4.1242745
214 0.51344422
215 -0.48587119
216 0.089730227
217 4.2202223
218 0.51540435
219 -0.4855087
220 0.13910035
221 4.1899477
222 0.51790597
223 -0.48570561
224 0.14609057
225 4.1694514
226 0.51777171
227 -0.48579511
228 0.072030912
229 4.1014511
230 0.50668223
231 -0.48596069
232 0.08274448
233 4.2034538
234 0.51758376
235 -0.48566533
236 0.084606153
237 4.1880861
238 0.51788359
239 -0.48575483
240 -0.046740218
241 4.2421148
242 0.51755243
243 -0.48555793
244 0.055262434
245 4.168986
246 0.51777171
247 -0.48589356
248 -0.020658901
249 4.1973989
250 0.5181163
251 -0.48581749
252 -0.15853008
253 3.9756898
254 0.48703622
255 -0.48634108
256 -0.030907051
257 4.2169644
258 0.51930222
259 -0.48573693
260 -0.53347362
261 -0.50785772
262 0.55011828
263 -0.50822021
264 -0.53347362
265 -0.50785772
266 0.55031518
267 -0.50822021
268 -0.53347362
269 -0.50785772
270 0.54990347
271 -0.50822021
272 -0.53347362
273 -0.50785772
274 0.55089248
275 -0.50822021
276 -0.53347362
277 -0.50785772
278 0.54966628
279 -0.50822021
280 -0.53393904
281 -0.50785772
282 0.55060607
283 -0.50821573
284 -0.53393904
285 -0.50785772
286 0.55193072
287 -0.50821573
288 -0.53393904
289 -0.50785772
290 0.55146978
291 -0.50821573
292 -0.53347362
293 -0.50785772
294 0.55146978
295 -0.50822021
296 -0.53347362
297 -0.50785772
298 0.55089248
299 -0.50822021
300 -0.53393904
301 -0.50785772
302 0.55151901
303 -0.50821573
304 -0.53347362
305 -0.50785772
306 0.55109834
307 -0.50822021
308 -0.5330082
309 -0.50785772
310 0.53611993
311 -0.50822021
312 -0.53347362
313 -0.50785772
314 0.55061502
315 -0.50822021
316 -0.53347362
317 -0.50785772
318 0.55109386
319 -0.50822021
320 -0.53347362
321 -0.50785772
322 0.55058817
323 -0.50822021
324 -0.53347362
325 -0.50785772
326 0.55048524
327 -0.50822021
328 -0.53347362
329 -0.50785772
330 0.5506911
331 -0.50822021
332 -0.53347362
333 -0.50785772
334 0.55104016
335 -0.50822021
336 -0.53347362
337 -0.50785772
338 0.55061054
339 -0.50822021
340 -0.53347362
341 -0.50785772
342 0.55108044
343 -0.50822021
344 -0.53347362
345 -0.50785772
346 0.55053894
347 -0.50822021
348 -0.53347362
349 -0.50785772
350 0.55054789
351 -0.50822021
352 -0.53347362
353 -0.50785772
354 0.55112967
355 -0.50822021
356 -0.53347362
357 -0.50785772
358 0.55163088
359 -0.50822021
360 -0.53347362
361 -0.50785772
362 0.55153243
363 -0.50822021
364 -0.53347362
365 -0.50785772
366 0.5519889
367 -0.50822021
368 -0.53347362
369 -0.50785772
370 0.55219476
371 -0.50822021
372 -0.53347362
373 -0.50785772
374 0.55067767
375 -0.50822021
376 -0.53347362
377 -0.50785772
378 0.55060607
379 -0.50822021
380 -0.53347362
381 -0.50785772
382 0.54987662
383 -0.50822021
384 -0.53347362
385 -0.50785772
386 0.5501138
387 -0.50822021
388 -0.53347362
389 -0.50785772
390 0.54670819
391 -0.50822021
392 -0.53347362
393 -0.50785772
394 0.54652024
395 -0.50822021
396 -0.53347362
397 -0.50785772
398 0.5494828
399 -0.50822021
};
\addplot [red, dashed, forget plot]
table {%
141 -0.53393904
141 4.2421148
};
\addplot [red, dashed, forget plot]
table {%
257 -0.53393904
257 4.2421148
};
\addplot [red, dashed, forget plot]
table {%
300 -0.53393904
300 4.2421148
};
\end{axis}

\end{tikzpicture}
\caption{Example 6. (Left) The zoom-in on the APTS-solution of Fig. \ref{fig_Ex6} illustrates that the breakpoint at $t=300$ was identified for representing an outlier that is slightly smaller than all other neighboring local minima. The level of neighboring local minima is emphasized by the dotted horizontal line. (Right)  The zoom-in shows that there are three data points around each local minimum. By setting $\gamma^\text{plat}=0.05$ the outlier is filtered out and the solution indicated in the bottom subplot of Fig. \ref{fig_Ex6} is obtained.}
\label{fig_Ex6_zoom}
\end{figure}
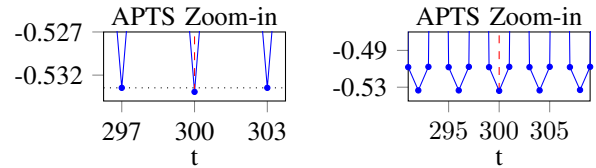

\begin{figure}
\centering
\input{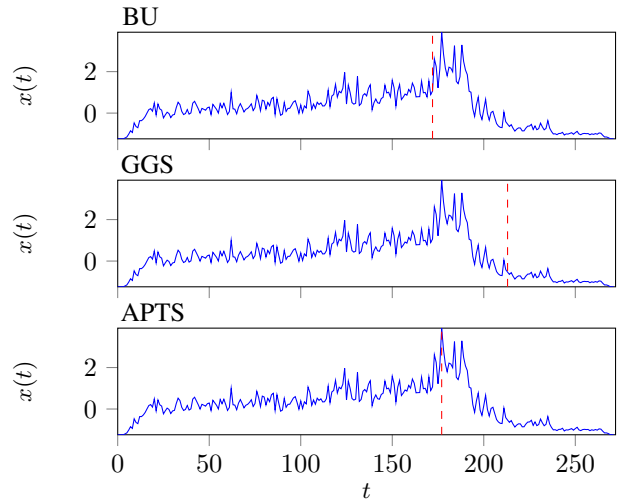}
\caption{Example 7. One-dimensional sample from dataset \textsf{'Lightning7'} \cite{dau2019ucr}. The breakpoint returned by APTS is exactly at the peak. BU does not set its breakpoint exactly at the peak. GGS segments two regions of different average amplitude levels.}
\label{fig_Ex7}
\end{figure}

\begin{figure}
\centering
\input{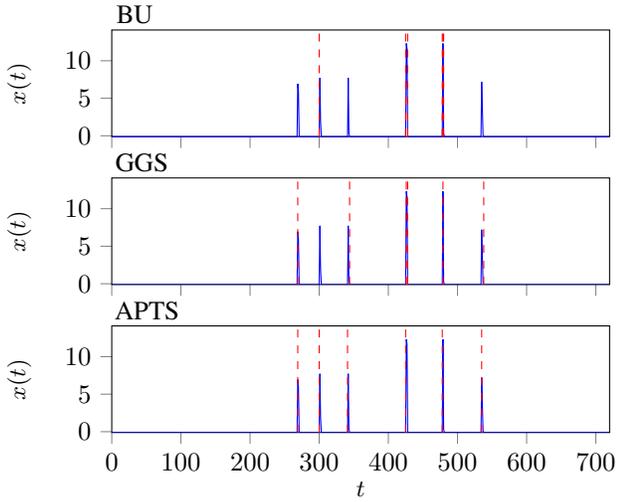}
\caption{Example 8. One-dimensional sample from dataset \textsf{'SmallKitchenAppliances'} \cite{dau2019ucr}. Only APTS identifies all six impulses correctly. In contrast, for both BU and GGS in at least one case two breakpoints are mapped closely to the same impulse.}
\label{fig_Ex8}
\end{figure}

\begin{figure}
\centering
\includegraphics[width=0.46\textwidth]{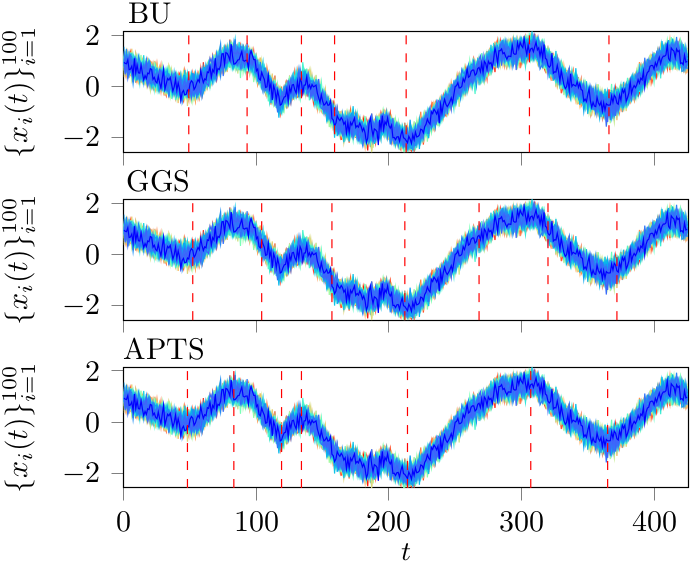}
\caption{Example 9. A one-dimensional sample from dataset \textsf{'Yoga'} \cite{dau2019ucr} is perturbed by zero-mean Gaussian noise with standard deviation of 0.2 to produce $n_x=100$ different time series. }
\label{fig_Ex9}
\end{figure}

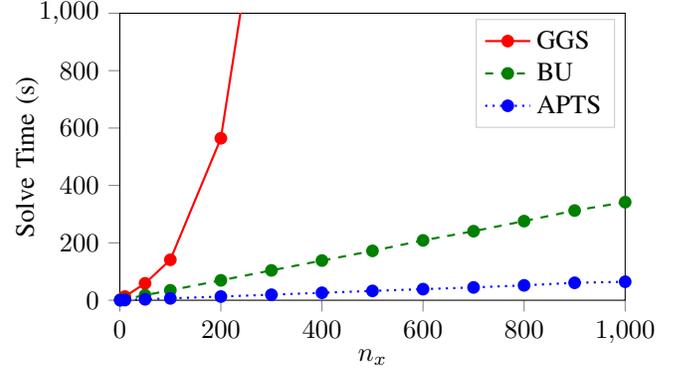
\begin{figure}
\centering
\begin{tikzpicture}

\begin{axis}[
xlabel={$n_x$},
ylabel={Solve Time (s)},
xmin=0, xmax=1000,
ymin=0, ymax=1000,
width=8.3cm,
height=5.4cm,
tick align=outside,
tick pos=left,
x grid style={lightgray!92.02614379084967!black},
y grid style={lightgray!92.02614379084967!black},
legend style={draw=white!80.0!black},
legend entries={{GGS},{BU},{APTS}},
legend cell align={left}
]
\addplot [red, solid, thick, mark=*, mark size=2, mark options={solid}]
table {%
1 1.89249205589
10 12.8149020672
50 58.9007060528
100 140.948530912
200 564.331665039
300 1690.01900601
400 3834.54837203
500 7452.8314209
};
\addplot [green!50.0!black, dashed, thick, mark=*, mark size=2, mark options={solid}]
table {%
1 0.630285978317
10 3.7547929287
50 17.5834369659
100 34.6318879128
200 69.3266961575
300 104.019581795
400 138.274966002
500 171.992558956
600 208.509974957
700 240.570383072
800 275.751890898
900 312.492367029
1000 341.70898509
};
\addplot [blue, dotted, thick, mark=*, mark size=2, mark options={solid}]
table {%
1 0.0802021026611
10 0.658583879471
50 3.19069886208
100 6.43537306786
200 12.9287478924
300 19.2645111084
400 25.9934110641
500 32.4742650986
600 38.6527719498
700 44.7052907944
800 51.8140451908
900 60.7592170238
1000 64.396627903
};
\end{axis}

\end{tikzpicture}
\caption{Large-scale Example 10. The full dataset \textsf{'HandOutlines'} \cite{dau2019ucr} has 1000 channels and 2709 data points per channel. While keeping $T=2709$ constant, BU, GGS and APTS were applied to subsets of the data with $n_x\in\{1,10,50,100,200,\dots,1000\}$, while recording the solve times. The y-axis is cut-off at 1000s solve time for clarity. See also Table \ref{tab_runtimes} for more detailed numerical results.}
\label{fig_CompTimesEx10}
\end{figure}

\begin{figure}
\centering
\includegraphics[width=0.48\textwidth]{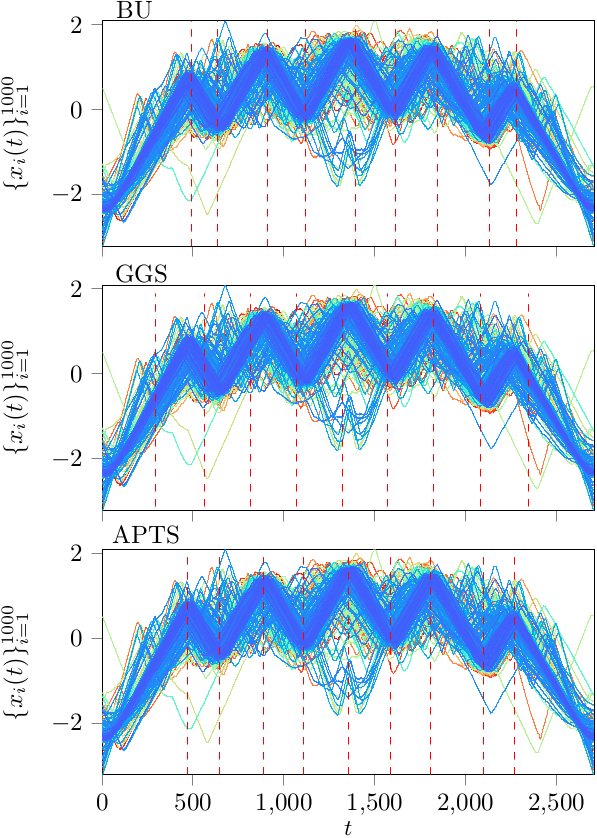}
\caption{Large-scale Example 10. Results for the full dataset \textsf{'HandOutlines'} \cite{dau2019ucr} are displayed. There are $n_x=1000$ channels with $T=2709$ data points for each channel.}
\label{fig_Ex10}
\end{figure}

All methods were implemented in Python. All experiments were run on an Intel i7-7700K CPU@4.2GHz$\times$8 processor with 15.6 GiB memory. Results for Examples 1-10 are visualized in Fig. \ref{fig_Ex1}-\ref{fig_Ex10}. Computational solve times are summarized in Table \ref{tab_runtimes}, and in Fig. \ref{fig_CompTimesEx10} for Example 10.

\setlength\figureheight{5cm}
\setlength\figurewidth{5cm}

\setlength\figureheight{5cm}
\setlength\figurewidth{5cm}

\setlength\figureheight{5cm}
\setlength\figurewidth{5cm}

\setlength\figureheight{5cm}
\setlength\figurewidth{5cm}
\begin{figure}
\centering
\input{Ex4.tex}
\caption{Example 4. One-dimensional sample from dataset \textsf{'EthanolLevel'} \cite{dau2019ucr}. Default APTS comes with $\gamma^\text{plat}=0$. The effect of $\gamma^\text{plat}=0.05$ is shown in the bottom subplot.}
\label{fig_Ex4}
\end{figure}

\setlength\figureheight{5cm}
\setlength\figurewidth{5cm}
\begin{figure}
\centering
\input{Ex5.tex}
\caption{Example 5. One-dimensional sample from dataset \textsf{'GunPointAgeSpan'} \cite{dau2019ucr}. Default APTS comes with $\gamma^\text{plat}=0$. The effect of $\gamma^\text{plat}=0.05$ is shown in the bottom subplot.}
\label{fig_Ex5}
\end{figure}

\setlength\figureheight{5cm}
\setlength\figurewidth{5cm}

\setlength\figureheight{5cm}
\setlength\figurewidth{5cm}

\setlength\figureheight{5cm}
\setlength\figurewidth{5cm}

\setlength\figureheight{5cm}
\setlength\figurewidth{5cm}

\setlength\figureheight{5cm}
\setlength\figurewidth{5cm}

Overall, APTS performs best and represents the most versatile approach for the handling of different shapes of time series data. While BU as expected appropriately segments the piecewise affine time series in Fig. \ref{fig_Ex1}, BU fails for the time series of Fig. \ref{fig_Ex2} with half-circle shapes. GGS is found to not be suitable for accurate identification of local maxima and minima. Instead, GGS seems to be best suited for data of varying  amplitude or noise levels such as in the examples of Fig. \ref{fig_Ex5}-\ref{fig_Ex6}. However, for $\gamma^\text{plat}>0$ these examples can be solved equally good by APTS and much faster. In fact, as Table \ref{tab_runtimes} shows, APTS offers consistently smaller solve times throughout all experiments in comparison to BU and GGS. Furthermore, APTS is much better scalable to high-dimensional data as Fig. \ref{fig_CompTimesEx10} demonstrates.

It is also found that the characteristic of APTS to return a solution for a prescribed upper bound $\hat{K}\leq K^\text{max}$, rather than for a scalar prescribing an exact desired number of segments, is not to be seen as a disadvantage. On the contrary, this characteristic of APTS permits to solve all 10 examples with a single set of hyperparameters, to tailor a suitable $\hat{K}$ for any given data, and thus to increase the degree of automation.

It is found that \emph{model-based} approaches such as BU \cite{keogh2001online} and GGS \cite{hallac2019greedy} are limiting, and not well generalizable to different shapes of data. For instance, the segmentation produced by GGS for the example in Fig. \ref{fig_Ex1} is not intuitive. Likewise, BU produces unsuitable results in Fig. \ref{fig_Ex2}. It therefore appears that a \emph{model-free} approach, such as APTS, is the most suitable choice to cover a large variety of different shapes of data. In a future extending hierarchical framework, on-top of a segmentation returned by APTS different models may be fitted to different segments.

For multivariate time series data the final segmentation returned by APTS is  equally valid over all channels. However, before reaching consensus separate segmentations are obtained separately for each channel. This property may be useful for extending upstream tasks such as time series clustering. This is pointed out since this is in contrast to GGS, where this property is absent and where only a combined segmentation over all channels is returned.

\begin{table}
\renewcommand{\arraystretch}{1.2}
\vspace{0.3cm}
\centering
\begin{tabular}{|c|cc|ccc|}
\hline
& & & \multicolumn{3}{c|}{Solve Times (s)}\\
 Ex. & $n_x$ & $T$  & BU \cite{keogh2001online} & GGS \cite{hallac2019greedy} & APTS \\ 
\hline
1 & 1 & 100 & 0.016 & 0.089  & \textbf{0.002} \\ 
2 & 1 & 100 & 0.017 & 0.116  & \textbf{0.002} \\
3 & 1 & 426 & 0.077 & 0.903  & \textbf{0.010} \\
4 & 1 &1751 & 0.369 & 2.293  & \textbf{0.040} \\
5 & 1&  150 & 0.026 & 0.161  & \textbf{0.003} \\
6 & 1 & 400 & 0.073 & 0.081  & \textbf{0.025}  \\
7 & 1 & 272 & 0.048 & 0.017  & \textbf{0.012}  \\
8 & 1&  720 & 0.135 & 0.405  & \textbf{0.039} \\
9 & 100 & 426 & 5.267 & 24.75  & \textbf{1.267} \\
10 &50 & 2709 & 17.58 & 58.90  & \textbf{3.19} \\ 
10 &500 & 2709 & 172.0 & 7452.8  & \textbf{32.47} \\ 
10 &1000 & 2709 & 341.7 & 65847.5 & \textbf{64.4} \\       
\hline
\end{tabular}
\caption{Comparison of solve times in seconds. The last three rows show results for the same Example 10, but for a different number of channels $n_x$. For GGS with $n_x=500$ and $n_x=1000$ solve times are equivalent to more than 2h and 18h, respectively. APTS can solve the full dimensional Example 10 with $n_x=1000$ in approximately the same time, 64.4s, as GGS requires for its solution with $n_x=50$. See also Fig. \ref{fig_Ex10}. Default APTS comes with $\gamma^\text{plat}=0$. The solve times for Examples 4-6 with $\gamma^\text{plat}=0.05$ were 0.249s, 0.003s, 0.059s, respectively.}
\label{tab_runtimes}
\end{table}

\section{Conclusion\label{sec_concl}}

A simple and scalable model-free algorithm for multivariate time series segmentation was presented. After a normalization step time series are treated channel-wise as surrogate stock prices that can be traded optimally a posteriori in a virtual portfolio holding either stock or cash. Linear transaction costs are interpreted as hyperparameters for noise filtering. The resulting trading signals as well as the resulting trading signals obtained on the reversed time series are used for unsupervised labeling, before a consensus over channels is reached that finally determines segmentation time instants. Proposed algorithm is called \emph{A posteriori Trading-inspired Segmentation} (APTS) and was compared to a popular bottom-up approach \cite{keogh2001online} fitting piecewise affine models and to a top-down approach fitting Gaussian models \cite{hallac2019greedy} on a variety of different synthetic data as well as real-world datasets from the UCR time series archive \cite{dau2019ucr}. Overall, APTS was found to be consistently faster while producing more intuitive segmentation results.

There are two main avenues for future work. First, the method may be extended for recursive online time series segmentation, e.g., within a moving horizon framework. Second, the method may serve as foundation for (i) further upstream time series analysis tasks including clustering, compression and forecasting \cite{plessen2020integrated} and (ii) for the organization of unstructured data lakes \cite{olawoyin2021open}.

%
%
\nocite{*}
\bibliographystyle{ieeetr}
\bibliography{myref}
%
%
%




\end{document}